%% file: __main_2025_haic.tex
\documentclass{article} %
\usepackage{iclr2025_conference,times}

\input{math_commands.tex}

\usepackage[utf8]{inputenc} %
\usepackage[T1]{fontenc}    %
\usepackage{hyperref}       %
\usepackage{url}            %
\usepackage{booktabs}       %
\usepackage{amsfonts}       %
\usepackage{nicefrac}       %
\usepackage{microtype}      %
\usepackage{graphicx}
\usepackage{setspace}
\usepackage{titlesec}
\usepackage[noabbrev,capitalize]{cleveref}
\usepackage{xspace}
\usepackage{enumitem}
\usepackage{comment}

\titlespacing*{\section}{0pt}{0.5ex plus 0ex minus 0ex}{0.25ex}
\titlespacing*{\subsection}{0pt}{0.5ex plus 0ex minus 0ex}{0.25ex}
\titlespacing*{\paragraph}{0pt}{0ex plus 0ex minus 0ex}{0.75em}

\newcommand{\ifprecedingtext}[1]{\ifvmode\relax\else#1\fi}

\renewcommand{\cite}{\citep}

\defcitealias{subcommittee2021}{Protecting Kids Online, 2021}

\title{Social Science Is Necessary for Operationalizing Socially Responsible Foundation Models}

\author{%
  Adam Davies\\
  Siebel School of Computing and Data Science\\
  The Grainger College of Engineering\\
  University of Illinois Urbana-Champaign\\
  \texttt{adavies4@illinois.edu}
  \And
  Elisa Nguyen\\
  T\"ubingen AI Center\\
  University of T\"ubingen
  \And
  Michael Simeone\\
  School of Complex Adaptive Systems\\
  Arizona State University
  \And
  Erik Johnston\\
  School for the Future of Innovation in Society\\
  Arizona State University
  \And
  Martin Gubri\\
  Parameter Lab
}

\iclrfinalcopy %
\begin{document}

\maketitle

\begin{abstract}
    With the rise of foundation models,
    there is growing concern about their potential social impacts.
    Social science has a long history of studying the social impacts of transformative technologies in terms of pre-existing systems of power and how these systems are disrupted or reinforced by new technologies.
    In this position paper, we build on prior work studying the social impacts of earlier technologies
    to propose a conceptual framework studying foundation models as sociotechnical systems, incorporating social science expertise to better understand how these models affect systems of power, anticipate the impacts of deploying these models in various applications, and study the effectiveness of technical interventions intended to mitigate social harms.
    We advocate for an interdisciplinary and collaborative research paradigm 
    between AI and social science across all stages of foundation model research and development to promote socially responsible research practices and use cases, and outline several strategies to facilitate such research.
\end{abstract}

\section{Introduction}

While the rapid recent development of generative foundation models is exciting for many potential applications (see, e.g., \citealt{touvron2023llama,jiang2023mistral,aryabumi2024aya,dubey2024llama}, etc.), important social impacts come along with rapid adoption, including worker displacement~\cite{leludec2023labour,casilli_waiting_2025,capraro2024impact}, use of copyrighted data for training models \cite{carlini_extracting_2021,somepalli_diffusion_2022,samuelson_science_2023,grynbaum_times_2023}, energy requirements and associated climate impact~\cite{tamburrini2022ai, luccioni2023counting}, and data privacy~\cite{carlini_extracting_2021,jo_lessons_2020,nasr_scalable_2023,kim_propile_2023}.
To develop socially responsible foundation models, we argue for \emph{proactive} consideration of such concerns across the whole research and development (R\&D) lifecycle from ideation to retirement of the technology. Anticipating social concerns can enable early discovery of unintended problems in the pipeline (e.g., biased data collection -- see \citealp{sambavisan2021everyone}) and inform interventions to mitigate undesired impacts \cite{hardt2016equality, Bommasani2021FoundationModels, sun2019mitigating}. 

Understanding and considering social impacts in the research and development of AI technology requires knowledge and experience studying complex social systems and interactions --
i.e., expertise in social science.
However, the domains of AI and social science research are largely siloed~\cite{selbst2019fairness,dahlin2021mind,sartori2022sociotechnical}, manifesting in differences in vocabulary~\cite{krafft2020defining}, publishing venues, and publishing practices  (e.g., the prestige of journals vs. conferences). %
Often, simplifying assumptions are made in AI research about social structures which may not hold in real life \cite{sartori2022sociotechnical} 
-- e.g., crowdsourcing annotators to align LLMs with so-called ``human values'' via reinforcement learning from human feedback (RLHF; \citealp{christiano2017deep,ouyang2022training}), despite the fact that such ``values'' are unique to the individual and may vary widely across cultures. 
To facilitate more socially responsible research, we advocate for an interdisciplinary paradigm integrating expertise in AI and social science %
throughout the technology lifecycle to anticipate and study potential social impacts of foundation models. %
First, we explore several relevant notions from social science to better contextualize how new technologies can impact society, highlighting how past failures to anticipate sociotechnical impacts have led to real social harms.
Building on these ideas, we propose a conceptual framework for integrating social science in foundation model research to understand social responsibilities, anticipate potential impacts, and develop technical innovations to create and deploy more socially responsible foundation models.
Finally, we consider incentives for tech firms and individual researchers that encourage or inhibit research and development toward socially responsible foundation models, and indicate several actionable suggestions to promote interdisciplinary collaboration between AI and social sciences through incentives, education, and skill development.

\section{Background}
\label{sec:background}

\paragraph{Social Systems of Power}\label{sec:bg_ss}

In one relevant intellectual tradition of social science, \emph{systems of power} -- the structures and institutions that shape, maintain, and distribute power within a society -- have been researched and described across a variety of theoretical paradigms and approaches including post-structuralism, socio-cultural theory, network analysis, and organizational theory \cite{linstead2003organization,roberts2012poststructuralism,martin2024sociology}. Scholars explore how institutions (such as governments, corporations, and social norms) distribute power and privilege, shaping social outcomes like prosperity, inequality, and marginalization. Many approaches starting toward the end of the 20th century emphasize the interconnectedness of race, class, gender, and other identities in understanding power dynamics \cite{crenshaw1991mapping,collins_black_1990}. These systems of power can be reproduced by new technologies such as social media platforms, recommendation algorithms, and search engines, which can amplify existing biases by encoding them into technological systems \cite{eubanks_automating_2018,noble_algorithms_2018,benjamin_2019_race}.

\paragraph{Technological Affordances}
In an oft-cited example of how racial systems of power can be codified in technology, \citet{noble_algorithms_2018} 
examines how Google Search in 2011-2013 reinforced longstanding harmful misrepresentations of Black women in, e.g., 
racist and sexist stereotypes that appeared in top autocompletions beginning with ``why are black women so'' versus ``why are white women so'', and the hyper-sexualization of Black women evidenced by the extreme prevalence of pornographic results when queried with ``black girls''.
We may understand such instances through the lens of \emph{technological affordances} -- i.e., the technology-mediated actions that are enabled, encouraged, or constrained by a technology with respect to an environment \cite{jones2020affordances}
-- in the specific context of information seeking \cite{zhao2020affordances,hirvonen2023affordances}, where the actions taken by web users (e.g., selecting an autocompletion or following a search result) are influenced by technologies that can implicitly reproduce existing systems of power (such as harmful stereotypes or sexual objectification), reciprocally shaping the digital information environment by driving search traffic and influencing users' beliefs to reinforce the social harms and inequities embedded in the technology \cite{vicente2023humans}.
In this work, we consider the technological affordances of foundation models, and the importance of social science for understanding how these affordances can reproduce or reshape existing systems of power.\footnote{
    Note that, while a primary focus of our work is the relevance of social science research to such considerations, the traditional subject matter and methods of humanities research are similarly critical \citep{klein2025provocations}.
}

\label{sec:bg_impact}

\paragraph{Social Media and Teen Mental Health}\label{sec:smu}
Before discussing foundation models, we first consider a more established technology where failing to take findings from social science and psychology into account has led to serious real-world harms: social media use (SMU) among teens, and its impact on their mental health.
Many studies have found a strong correlation between SMU and 
diagnoses of mood and body-image disorders
\cite{barry2017adolescent,gupta2022reviewing,costello2023algorithms,weigle2024social};
and while it is difficult to establish a direct causal relationship,
the limited causal evidence available suggests that SMU is indeed an important contributor to these negative impacts \cite{bozzola2022use,weigle2024social}.
One possible solution that has been proposed to help mitigate such harms is to redesign content recommendation feeds to de-prioritize engagement metrics, as
there is clear evidence that recommender systems optimized for user engagement suggest harmful content at a far higher rate than systems that do not \cite{banker2019algorithm}.
For instace, despite early internal user studies conducted at Facebook and Instagram finding that simple adjustments to engagement-based algorithmic design choices could indeed mitigate negative impacts on teen mental health \cite{wells2021facebook,hao2021facebook}, the teams conducting this research were shuttered and the corresponding changes were never adopted at scale because they also led to lower advertising revenues (\citealp{hao2021facebook, mac2021whistleblower,costello2023algorithms}; \citetalias{subcommittee2021}).

\begin{figure}[t]
    \centering
    \includegraphics[width=\linewidth]{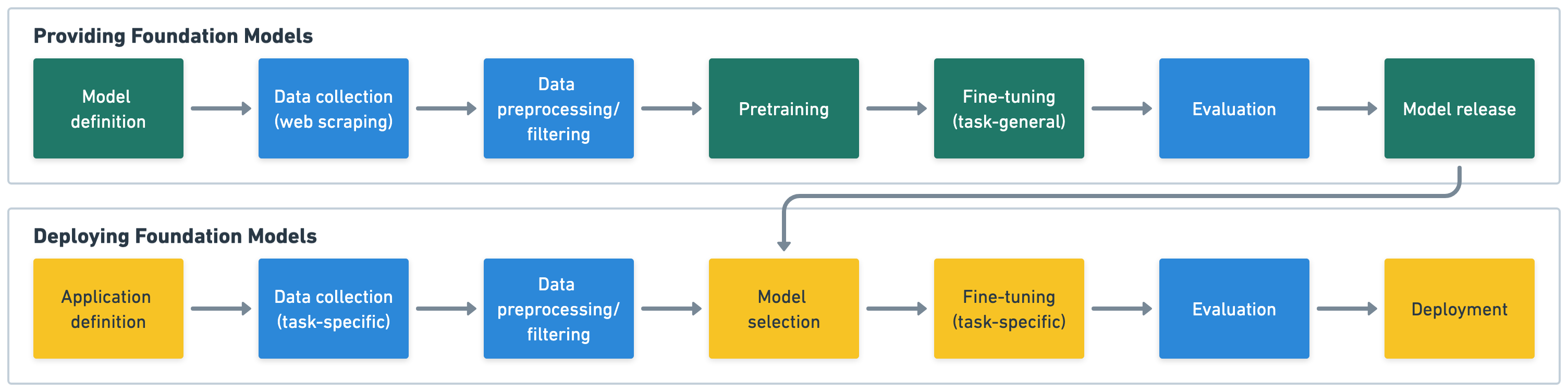}
    \caption{\textbf{Steps of the Foundation Model R\&D pipeline.} 
    The top pipeline illustrates the stages for training a foundation model (providers), while the bottom pipeline describes the stages of deploying foundation models (deployers).}
    \label{fig:pipeline}
    \vspace{-1.5em}
\end{figure}

\paragraph{The Foundation Model Pipeline}
Throughout this work, we will focus on a more recent, potentially socially-transformative technology, \emph{foundation models} (i.e., self-supervised deep learning models trained on large-scale web data, such as LLMs). 
The process of creating or deploying foundation models can be visualized as a pipeline representing the different stages of the research and development (R\&D) process,\footnote{
    Throughout this work, we use foundation model \emph{research and development (R\&D)} very broadly, where ``research'' is intended to cover all aspects of foundation model research -- e.g., from basic research involving model architectures, loss functions, fine-tuning paradigms, etc., all the way to benchmarking existing models or developing applied techniques to improve models' performance for specific tasks.
}  as visualized in Figure~\ref{fig:pipeline}, where decisions in each step of the pipeline can carry important consequences for later stages -- for example, sub-optimal data collection and filtering choices can have serious implications for downstream model robustness and lead to preventable social harms ~\cite{sambavisan2021everyone}. 
Following the EU AI Act~\cite{aiact}, we categorize model \emph{providers} as those who develop a general-purpose AI model; and model \emph{deployers} as those who develop a product or service leveraging such models for a specific use case where providers can also be deployers of their own models (e.g., OpenAI is the provider of ChatGPT, and a company that calls the ChatGPT API in a user-facing product would be a deployer).

\section{Operationalizing Socially Responsible Foundation Models}

To provide and deploy foundation models in a socially-responsible manner, we argue that it is necessary to involve social science expertise throughout the foundation model R\&D process.
In particular, we propose a conceptual framework to decompose this task into three key components:
\begin{enumerate}[itemsep=1pt,topsep=0pt, partopsep=0pt,leftmargin=16pt]
    \item \textbf{Understanding systems of power:} How do disruptive technologies like foundation models reproduce or reshape existing systems of power? 
    What affordances would best promote desirable effects on these systems?
    \item \textbf{Designing technical interventions:} 
    How can the foundation model R\&D pipeline be modified to align models with target affordances?
    \item \textbf{Anticipating social impacts:} 
    What social impacts may result from deploying a model with the target affordances in a specific context?
\end{enumerate}

While AI researchers are well-positioned to 
study technical interventions (2),
this is an entirely different question from 
understanding systems of power (1) or anticipating social impacts (3), which are better suited to social scientists.
However, there is still a role in each component for AI research, as it is nonetheless important to provide robust, quantifiable, and computationally tractable definitions of 
desired foundation model affordances (e.g., it is necessary to specify affordances in terms that can be learned by models in encouraging socially representative model outputs, prohibiting the use of models for generating toxic content, etc.),
as well as to carry out systematic empirical evaluations of corresponding model behaviors to predict alignment with the intended affordances, which are both tasks where AI expertise is essential.
As such, 
interdisciplinary collaboration between AI and social science is required to address the challenges associated with each of these components.
Specifically, we argue that it is critical to involve social science in foundation model research, development, and deployment
in order to 
(1)~\emph{proactively consider interactions between foundation model affordances and sociotechnical systems of power,} 
and (2)~\emph{anticipate the impacts associated with deploying these models in a given context,} as explored below.

\paragraph{Responsible Model Providers Proactively Consider Systems of Power.}\label{sec:providers}
Large scale, web-scraped data is an essential igredient for training all foundation models; and such data is shaped by sociotechnical systems of powers in subtle, complex, and systematic ways.
For instance, Wikipedia, which has been heavily relied upon as a large and high-quality knowledge resource in many LLM training datasets \cite{touvron2023llama,gao2021thepile,soldaini2024dolma}, underrepresents women and non-binary figures \cite{Graells2015first,hube2017wikibiais,falenska2021assessing,Tripodi2023wikipedia,FerranFerrer2023WikipediaGG} -- e.g., only 19\% of biographies are about women \cite{Tripodi2023wikipedia}. This \textit{Wikipedia gender gap} is well studied in social science (see \citealt{FerranFerrer2023WikipediaGG} for a comprehensive survey on the topic) as a complex systemic phenomenon. Using the conceptual framework of fields of visibility, \citet{beytia_visibility_2022} analyze content asymmetries on Wikipedia as a system composed of diverse agents affecting content in terms of representation, characterization, and structural placement. 
For model providers to avoid reinforcing systematic under- and mis-representation, it is important to be aware of such phenomena and act to mitigate resulting bias (e.g., by actively collecting under-represented data \citealp{jo2020archives}; or implementing debiasing techniques  \citealp{mehrabi2021survey,parraga2025fairness}). %
However, such techniques are not a ``silver bullet'' solution \cite{anwar2024foundational}, given the wide variety of statistical notions of bias that can be contradictory and entail tradeoffs \cite{verma2018fairness,carey2023statistical} and the problematic simplifying assumptions required to mitigate biased representation by way of statistical methods \cite{bode2020you}.
Thus, it is a key responsibility of model providers to study and transparently communicate learned biases to downstream model deployers,
as understanding and documenting the sources and effects of potential biases can provide the necessary context for selecting the application areas of a model \cite{sherman2024power, klein2025provocations}. For example, \citet{mitchell2019card} argue that models should be distributed alongside \textit{model cards} that report metrics at a disaggregated level for cultural, demographic, or phenotypic population groups,\footnote{
    E.g., model providers can report intended uses and potential limitations using Hugging Face's \href{https://huggingface.co/docs/hub/en/model-cards}{model and dataset card features}, inspired by \cite{mitchell2019card}.
} and \citet{klein2025provocations} further advocate for detailed documentation of data collection and/or generation procedures.

To illustrate the importance of considering systems of power for all stages in the foundation model pipeline,
consider a scenario where 
these systems
are not taken into account by model providers.
Here, whatever systemic inequities are present in the model's training corpus (e.g., under-representation of women, harmful stereotypes of racial or ethnic minorities, etc.) can easily be learned and reproduced by the model, naturally affording corresponding harmful use cases \cite{sambavisan2021everyone,weidinger2021ethical}.
Despite the common counter-argument that web-scraped data simply reflects the reality of what content appears on the web, and that 
it is not the responsibility of model providers to mitigate any given notion of bias in one's pre-training corpus (as highlighted by \citealt{birhane2023hate}),
the alternative \emph{laissez faire} approach, where systems of power are not taken into account whatsoever, 
can lead to an avoidable ``race-to-the-bottom'' collective action problem among model providers, deployers, and end-users.
In this case, each deployer utilizing the provided model would need to decide whether and how to account for social risks or harms on their own, 
and those who make the greatest effort to mitigate them
will incur a greater time and cost in doing so relative to less scrupulous competitors.
That is, where many deployers might \emph{prefer} that bias had been better mitigated by providers, it may not be possible for them to take on this task on their own while maintaining competitiveness; and ethics-minded employees may be disempowered to take collective action \cite{nedzhvetskaya2024role} due to financial precarity, immigration status, workplace culture, or organizational incentives \cite{widder2023s}.
Similarly, from a user's perspective, risks and harms might only be addressed (if at all) after some level of harm has already been done (given the competitive disadvantage associated with anticipating and proactively addressing possible harms); 
and providers will face a lack of trust in the safety of their models on the part of deployers and end users \cite{keymolen2024trustworthy}.
In such cases, we argue that there should be a \emph{duty of care} \cite{witting2005duty,arbour2008responsibility,welsh2012should} to anticipate, transparently communicate, and act to mitigate the propagation of discriminatory (or otherwise harmful) systems of power to avoid the social dilemma described above. We further consider pragmatic motivations for model providers to address such concerns in \cref{sec:techincentives}.

\paragraph{Responsible Model Deployers Address Application-Specific Social Impacts.}\label{sec:deployers}

In deploying existing foundation models for a specific application context, model deployers are best placed to consult social scientists in (a) anticipating potential social impacts associated with their specific intended application, and (b) designing and studying effective affordances,
where various foundation model applications require expertise from different disciplines in social science. For example, consider an application leveraging foundation models to edit photos before they are posted to social media.
Many popular social media platforms already afford users to edit selfies using ``beauty filters'' \cite{eshiet2020beauty,ryan2021beauty} 
that modify their appearance to align with a socially-constructed representation of conventional attractiveness or high social status \cite{javornik2022lies,burnell2022snapchat}.
If we consider societies where such a notion includes being thin, then these filters are expected to reduce the apparent weight of users in photos \citep{eshiet2020beauty,ateq2024association}; 
or in the context of societies where this notion is associated with lighter skin tone, these filters have been shown lighten the apparent skin tone of users \citep{riccio2024mirror, trammel2023artificial}. %
In the former case, the filter affordance reinforces a culture of ``fatphobia'', stigmatizing heavier individuals and creating unrealistic body standards \cite{robinson1993fat}; and in the latter case reinforces racial caste systems, such as White supremacy \cite{bonilla2001white}.
Indeed, filter affordances predating the era of generative foundation models have already been implicated in teen body image disorders \cite{burnell2022snapchat, tremblay2021filters}, and it is reasonable to expect that more powerful generative models will potentially lead to further such harms.
As such, just as in the case explored in \cref{sec:bg_impact} on the relationship between social media use and teen mental health,
social psychologists should likewise be consulted in this case to anticipate the potential impacts of foundation model-enabled filters, and in studying the effectiveness of possible interventions to mitigate harmful affordances.

Note that, while we have focused here on social media platform affordances enabled by foundation models, analogous arguments can be made for many other aspects of society and require expertise from different disciplines in social science.
For instance, in the workplace, foundation model-enabled affordances are predicted to carry widely varying net impacts on wages and labor markets depending on the speed and manner in which various workplace tasks are automated \cite{acemoglu2024tasks,acemoglu2024learning}; in education, they are expected to help democratize education worldwide while also leading to broader and more systemic bias in educational assessment and college admissions \cite{akgun2022artificial,baker2022algorithmic} or exacerbating the digital divide \cite{capraro2024impact,mannekote2024large}; and so on.
Each of these considerations requires consultation with relevant domain-area experts to anticipate potential impacts and design mitigation strategies.

\section{Facilitating Socially Responsible Foundation Models}
\label{sec:looking-forward}

Despite the rationale and approach for researching and developing more socially-responsible foundation models articulated above, it is unrealistic to expect all stakeholders to opt for such an approach on ethical merit alone, as doing so may conflict with other incentives such as short-term profits. Below, we consider incentives for (1) tech firms providing and deploying foundation models, and (2) interdisciplinary AI + social science research, and suggest potential interventions that may aid in (re)structuring incentives in favor of interdisciplinary work toward more socially-responsible foundation models.

\paragraph{Incentives for Tech Firms}\label{sec:techincentives}
Failure to anticipate and proactively address deleterious effects of social media use on teen mental health, as discussed in \cref{sec:smu}, has resulted in substantial brand harm and increased regulatory oversight for social media companies \cite{wells2021facebook,hao2021facebook,costello2023algorithms}.
In contrast, there is evidence to suggest that tech firms may benefit financially from prioritizing social responsibility in providing and deploying foundation models.
As outlined by \citet{gillan_firms_2021}, there is a large and growing body of research in financial economics suggesting that more socially-responsible firms tend to see superior financial performance and stability in the long term -- specifically, that the Environmental, Social, and Governance (ESG) and the Corporate Social Responsibility (CSR) profiles of firms are strongly related to lower risk, higher performance, and higher value. For example, \citet{lins_social_2016} show that high-CSR firms had better stock returns, profitability, growth, and sales per employee, compared to low-CSR firms during the 2008--2009 financial crisis, suggesting that investments in social capital can pay off in times of economic crisis. Furthermore, \citet{hong_crime_2019} estimate that, in the aggregate, high-ESG firms face 65\% lower sanctions from prosecutors. 
Thus, we hypothesize that tech firms prioritizing social responsibility in providing and deploying foundation models may observe similar financial benefits.

\paragraph{Incentives for Interdisciplinary Collaboration}
Interdisciplinary collaboration between social science and AI research runs contrary to some key incentives for researchers' career advancement. For instance, interdisciplinary publication venues are unlikely to be among the top-tier venues in each respective discipline \cite{campbell2005overcoming}; and while most top social science venues are journals, most top AI venues are conferences instead.
As such, even the best interdisciplinary work is less likely to be adequately recognized, awarded, cited, and disseminated \cite{pellmar2000barriers}.
Similar issues exist for other key factors in academic career advancement beyond publication venues, such as grant review \cite{lindell_bromham__2016} and degree requirements for university students \cite{amelink_transdisciplinary_2024}.
The following is a preliminary list of suggestions for attenuating the
cost of interdisciplinary collaboration, though it is not intended to be exhaustive:
\begin{itemize}[itemsep=1pt,topsep=0pt, partopsep=0pt,leftmargin=16pt] 
    \item 
    As in the FAccT conference,\footnote{
        See \url{https://facctconference.org/2024/cfp}.
    } more AI conferences could offer the optional choice of non-archival paper submissions (in addition to the standard archival submission), allowing researchers from other fields to later submit their conference papers to discipline-specific journals.
    \item 
    Research institutions could better consider interdisciplinary work in career advancement \cite{pellmar2000barriers} and funding proposals assessment \cite{lindell_bromham__2016}, and offer specialized funding opportunities and sabbaticals, allowing researchers to explore new ideas and collaborations in a wider context~\cite{ioppolo_how_2023}.
    \item Existing practices for promoting socially-responsible research can be further promoted by publication venues (e.g., by making model and dataset cards \citealp{mitchell2019card} a mandatory part of certain submission types) to expand their adoption as a standard in the research community.
    \item Promoting interdisciplinary education helps provide the next generation of researchers with the foundations to integrate methods from, and facilitate collaborations with, fields beyond their primary research area. For instance, awarding degree credit for courses from other disciplines encourages students to learn the essentials of these fields \cite{amelink_transdisciplinary_2024}, 
    and embedding ethics education in technical courses can improve students' abilities to engage in relevant ethical discussions \cite{horton2022embedding}. 
    \item Researchers considering a more interdisciplinary agenda could broaden their expertise with workshops, tutorials, or short courses provided by researchers from other fields. For example, we suggest that AI and social science conferences open tutorial calls to researchers outside their respective disciplines.
\end{itemize}

\section{Conclusion}

In this work, we have advocated for interdisciplinary research between AI and social science in the context of foundation models like LLMs, focusing on the importance of social science in understanding the affordances and social impacts of such transformative technologies. We outlined the importance of interdisciplinary expertise and collaboration throughout the foundation model R\&D pipeline, highlighted the associated responsibilities and benefits for model providers and deployers, and provided actionable suggestions to promote collaboration between AI and social science.
Finally, we discuss a few important considerations for future work in \cref{sec:limitsandfuturework}.

\section*{Acknowledgements}
We thank Alan Craig, Dave Buckley, and Linda Derhak
for their help in facilitating this project and sharing valuable feedback.
This work is supported in part by the National Science Foundation and the Institute of Education Sciences, U.S. Department of Education, through Award \#2229612 (National AI Institute for Inclusive Intelligent Technologies for Education). Any opinions, findings, and conclusions or recommendations expressed in this material are those of the author(s) and do not necessarily reflect the views of National Science Foundation or the U.S. Department of Education.
This material is based upon work supported in part by the National Science Foundation under Grant No. 2217706.
The authors thank the International Max Planck Research School for Intelligent Systems (IMPRS-IS) for supporting Elisa Nguyen. 
This work was supported in part by the T\"ubingen AI Center and the Parameter Lab company.

\bibliography{_references}
\bibliographystyle{iclr2025_conference}

\appendix
\section{Future Work}\label{sec:limitsandfuturework}

An important consideration regarding the interventions suggested in \cref{sec:looking-forward} is that interdisciplinary research can be expensive and time-consuming \cite{pellmar2000barriers}, and bringing in diverse perspectives always carries the potential to dilute research focus with competing visions and priorities \cite{committee2005facilitating}.
We suggest that future work could consider performing more comprehensive cost-benefit analyses 
along multiple dimensions (cf. \citealp{slaper2011triple})
to assess the resources needed to achieve the benefits outlined above, making it possible to more effectively manage these research tradeoffs.
More broadly, we recommend that AI experts and labs researching, developing, or deploying foundation models reflect on incorporating interdisciplinary collaboration within their team and their research topic more broadly, particularly in promoting socially responsible affordances and studying potential social impacts of their work.
Neither AI nor social science holds all the answers regarding how to develop safe, beneficial, and socially responsible foundation models; and it is critical that both disciplines work more closely together toward this goal, rather than ``siloing'' research for such a potentially transformative technology.

\end{document}

%% file: math_commands.tex
\usepackage{amsmath,amsfonts,bm}

\def\eqref#1{equation~\ref{#1}}

\def\1{\bm{1}}

\DeclareMathAlphabet{\mathsfit}{\encodingdefault}{\sfdefault}{m}{sl}
\SetMathAlphabet{\mathsfit}{bold}{\encodingdefault}{\sfdefault}{bx}{n}